\documentclass{article}

\usepackage{PRIMEarxiv}

\usepackage[utf8]{inputenc} % allow utf-8 input
\usepackage[T1]{fontenc}    % use 8-bit T1 fonts
\usepackage{hyperref}       % hyperlinks
\usepackage{url}            % simple URL typesetting
\usepackage{booktabs}       % professional-quality tables
\usepackage{amsfonts}       % blackboard math symbols
\usepackage{nicefrac}       % compact symbols for 1/2, etc.
\usepackage{microtype}      % microtypography
\usepackage{lipsum}
\usepackage{multirow}
\usepackage{fancyhdr}       % header
\usepackage{graphicx}       % graphics
\graphicspath{{media/}}     % organize your images and other figures under media/ folder

\title{Predicting Patient Self-reported Race From Skin Histological Images with Deep Learning
\thanks{\textit{\underline{Citation}}: 
\textbf{Accepted to the MICCAI Workshop on Fairness of AI in Medical Imaging (FAIMI), 2025. Full citation will be provided after final publication.}} 
}

\author{
  Shengjia Chen, Ruchika Verma, Jannes Jegminat, Eugenia Alleva, Thomas Fuchs, Gabriele Campanella$^{*}$ \\
  Windreich Department of Artificial Intelligence and Human Health, \\
  Icahn School of Medicine at Mount Sinai, New York, USA \\
  Hasso Plattner Institute for Digital Health at Mount Sinai, \\
  Icahn School of Medicine at Mount Sinai, New York, USA \\
  \texttt{Corresponding author:gabriele.campanella@mssm.edu} \\
  \And
  Kuan-lin Huang \\
  Windreich Department of Artificial Intelligence and Human Health, \\
  Icahn School of Medicine at Mount Sinai, New York, USA \\
  Mount Sinai Center for Transformative Disease Modeling, \\
  Icahn School of Medicine at Mount Sinai, New York, USA \\
  \And
  Kevin Clare, Brandon Veremis \\
  Department of Pathology, Molecular and Cell-Based Medicine, \\
  Mount Sinai Health System, New York, USA \\
}

\begin{document}
\maketitle

\begin{abstract}
Artificial Intelligence (AI) has demonstrated success in computational pathology (CPath) for disease detection, biomarker classification, and prognosis prediction. However, its potential to learn unintended demographic biases, particularly those related to social determinants of health, remains understudied. This study investigates whether deep learning models can predict self-reported race from digitized dermatopathology slides and identifies potential morphological shortcuts. Using a multisite dataset with a racially diverse population, we apply an attention-based mechanism to uncover race-associated morphological features. After evaluating three dataset curation strategies to control for confounding factors, the final experiment showed that White and Black demographic groups retained high prediction performance (AUC: 0.799, 0.762), while overall performance dropped to 0.663. Attention analysis revealed the epidermis as a key predictive feature, with significant performance declines when these regions were removed. These findings highlight the need for careful data curation and bias mitigation to ensure equitable AI deployment in pathology. Code available at: \url{https://github.com/sinai-computational-pathology/CPath_SAIF}.
\end{abstract}

% keywords can be removed
\keywords{Computational Pathology \and AI Fairness \and Dermatopathology}

\section{Introduction}
Bias and disparities in Machine Learning (ML)-based biomedical and healthcare applications have been widely studied by stratifying model performance across demographic groups \cite{seyyed2021underdiagnosis}. Recent studies have shown that ML models can propagate or even exacerbate existing healthcare inequalities due to dataset bias, arising from differences in disease prevalence, clinical presentation, and annotation inconsistencies across demographic groups \cite{nazer2023bias,jones2024causal}. While algorithmic fairness techniques have been explored to mitigate bias \cite{yang2023adversarial}, several studies have also demonstrated that feature–confounder correlations, such as the presence of treatment artifacts or institution-specific markers, can undermine model generalizability and fairness \cite{oakden2020hidden,howard2021impact,hill2024risk}.

In computational pathology (CPath), deep learning (DL) models have shown promise in disease detection \cite{hosseini2024computational}, biomarker classification \cite{el2025whole}, and prognosis prediction \cite{song2023artificial}, but demographic disparities in performance have also been reported in recent studies\cite{vaidya2024demographic}. While biases and demographic shortcuts are well-studied in medical imaging \cite{yang2024limits}, particularly radiology \cite{gichoya2022ai,adleberg2022predicting}, similar investigations in histopathology remain limited. Histological slides capture complex tissue morphology, cellular structures, and microenvironmental characteristics, but it is unclear whether these reflect demographic variations. Identifying such associations is crucial to understand confounders that may influence differential model performance in CPath \cite{williams2023skin}.

In this study, we investigate whether the DL models can infer self-reported race from histological images using skin histology data collected across multiple sites within a health system, without specific curation. Skin histology provides a unique opportunity for this analysis, as characteristics related to melanin and pigmentation—while visibly distinct in clinical dermatology \cite{harvey2024integrating}—are not readily apparent in histological images, making it unclear whether DL models can still capture race-associated patterns. By focusing on a single organ system, we effectively control for potential confounding variables that would present greater challenges in a more heterogeneous dataset. Using widely validated tile-level foundation models (FMs) in CPath \cite{campanella2024clinical} combined with explainable attention-based model AB-MIL \cite{ilse2018attention,chen2024benchmarking}, we examine whether tissue and cellular features can predict self-reported race. Furthermore, we implement a histomorphological phenotype learning framework \cite{claudio2024mapping} to identify morphologies associated with high attention regions, providing biological insights into model behavior.

\section{Related Work}
Recent studies have shown that DL models can predict self-reported race with high accuracy across medical imaging modalities, particularly in radiology \cite{zou2023implications}. Adleberg et al. \cite{adleberg2022predicting} reported an AUC of 0.911 for race prediction using chest radiographs, a capability that persists across modalities even when undetectable to human experts \cite{gichoya2022ai}. Beyond classification, race-related feature encodings have been observed in chest X-ray foundation models \cite{glocker2023risk} and brain age prediction models trained on MRI, with both showing performance disparities and statistically significant distribution shifts across demographic subgroups \cite{piccarra2023analysing}. In histopathology, stain variability and site-specific digital signatures can correlate with ethnicity and inflate model performance \cite{howard2021impact,kheiri2024bias}. Additionally, models trained for diagnostic tasks can encode racial information, with diagnostic accuracy positively associated with race prediction performance, even after mitigation efforts \cite{vaidya2024demographic}. Extending these findings to CPath, our work investigates histomorphological features associated with self-reported race in dermatopathology, aiming to identify potential biological confounders and assess their influence on model predictions.

\section{Methods}

\subsection{Dataset}
Self-reported race, a social construct with known correlations to differential health outcomes and a widely recognized social determinant of health, was collected from patient records and questionnaires at Mount Sinai health system. Patients with self-reported race equal to "unknown" or "not reported" were removed. Our private dataset consists of digitized slides from all available skin specimens, assembled from multiple sites in New York city, with all slides scanned on a Philips Ultrafast scanner. The dataset exhibits a diverse racial distribution with the overall patient population at Mount Sinai health system, closely matching the city's demographics. Although the White group is slightly overrepresented (39.3\%), this imbalance is relatively minor compared to other widely used histological datasets, such as TCGA, where 73.7\% of samples are from White patients. Self-reported race data are provided for comparison in Table \ref{tab:dataset_summary}. 

The dataset was generated from all available dermatopathology specimens within the health system, rather than being curated for a specific disease or prediction task. This includes a wide range of skin conditions such as hemorrhoids, melanoma, basal cell carcinoma (BCC), seborrheic keratosis, squamous cell carcinoma (SCC), and various types of inflammatory and infectious dermatoses. Additionally, the potential site-specific signature, as suggested in \cite{howard2021impact}, has been controlled for since all slides collected from different sites were stained and digitized in a central laboratory within the health system.

\begin{table*}[!ht]
\centering
\caption{Summary of the skin dataset by self-reported race compared with Mount Sinai healthcare system, New York city population, and public source (TCGA).}
\label{tab:dataset_summary}
\scriptsize
\begin{tabular}{lccccc}
\toprule
Self-reported & \multicolumn{2}{c}{Skin Cohort} & Health System & City & TCGA  \\
\cmidrule(lr){2-3} \cmidrule(lr){4-6} 
Race               & \# Slides (\%) & Patients \% &   & Population \% &  \\
\midrule
White              & 2,151 (40.8\%)         & 39.3      & 43.1      & 31.2 & 73.7 \\
Black              & 1,015 (19.3\%)         & 19.0        & 21.7      & 29.9 & 10.3 \\
Hispanic/Latino    & 868 (16.5\%)           & 16.8       & 18.5      & 21.0   & 8.5 \\
Asian              & 687 (13.1\%)           & 15.7       & 10.3      & 5.7 & 7.1 \\
Other             & 543 (10.3\%)           & 9.3       & 6.4       & 4.5  & 1.8 \\
\midrule
Total Number             & 5,266                  & 2,471     & 114,947   & 8M   & 23,276 \\
\bottomrule
\end{tabular}
\end{table*}

\subsection{Experimental Setup} 
To investigate the capability of DL models to classify self-reported race from histological images, we implemented a classification pipeline leveraging FM for feature extraction. Each slide was assigned a label corresponding to the self-reported race of the patient. The dataset was split 80/20 for training and validation at the patient level, with no separate test set allocated since generalization was not the focus. Tissue tiles were extracted at 20x magnification, and tile-level embeddings were generated using four pretrained FMs: SP22M \cite{campanella2023computational}, UNI \cite{chen2024towards}, GigaPath \cite{xu2024whole}, and Virchow \cite{vorontsov2024foundation}, followed by an attention-based MIL (AB-MIL) model \cite{ilse2018attention} for slide-level aggregation.  We selected AB-MIL due to its ability to efficiently learn informative tile-level attention scores that highlight discriminative regions within a slide while maintaining interpretability. The attention mechanism allows the model to quantitatively assess each tile’s contribution to the slide-level racial prediction. These models are specified by the count of skin slides utilized in their pretraining and their model size: SP22M (1426 slides, 22M) \cite{campanella2023computational}, UNI (3653 slides, 303M) \cite{chen2024towards}, GigaPath (2243 slides, 1135M) \cite{xu2024whole}, and Virchow (273,893 slides, 1,488M) \cite{vorontsov2024foundation}. 

During training, a weighted loss function was applied to ensure class balance, and models were trained for 40 epochs using the AdamW optimizer \cite{loshchilov2017fixing} with an initial learning rate of 0.0005, a 5-epoch warm-up, and a cosine decay schedule. A batch size of 512 was employed, and the final model checkpoint was evaluated on the validation set for performance and attention analysis. To ensure reproducibility and stability, Xavier initialization \cite{glorot2010understanding} was applied with three fixed random seeds (0, 42, 2025), and output probabilities and attention scores were averaged across these runs. All training was conducted on a single H100 GPU.

\subsection{UMAP Visualization}
To better understand the histological patterns that are important for the self-reported race prediction task, we utilized a histomorphological phenotype learning framework \cite{claudio2024mapping} to efficiently analyze regions of high attention. This tool also enables the efficient segmentation of tile-level histological structures, allowing us to study the relative attention given to different tissue compartments. Instead of training a segmentation model from scratch, we leveraged SP22M \cite{campanella2023computational} to extract tile features, which were then projected into a 2D UMAP space \cite{mcinnes2018umap} for visualization. Pathologists annotated a few landmark tiles to identify key tissue structures, allowing us to locate similar tiles in the UMAP space. Through iterative refinement, a Random Forest classifier was trained to segment regions of interest (ROI) based on UMAP-embedded tile features, defining ROIs as areas where at least 20\% of pixels corresponded to a given morphological class. Representative morphological classes identified in the UMAP space included epidermis, inflammation, gastrointestinal (GI) tissue, bone, adipose tissue (fat), blood, smooth muscle, skeletal muscle, ducts, and oncocytes, as well as common artifacts such as ink, cautery, and coverslip edges. Two pathologists validated these annotations before proceeding with stratified attention analysis.

\subsection{Attention Scores and Distribution Analysis}
The attention score for each self-reported race class was obtained from AB-MIL \cite{ilse2018attention}, incorporating a multi-head mechanism similar to CLAM \cite{lu2021data} to output distinct attention scores for each race groups. Each tile within a slide received an attention score corresponding to the race prediction head, indicating its contribution to the model's classification decision. To enable cross-slide comparisons, attention scores were normalized across all tiles in the validation dataset to a [0,1] range. We then compared mean attention scores between ROI and non-ROI areas to assess the relationship between attention and tissue morphology, investigating whether specific tissue types contributed more significantly to the model's decision-making process.

\section{Results}
To evaluate the potential bias of disease distribution on race prediction, we curated three versions of datasets. Model performance was evaluated using one-vs-rest (OvR) area under the curve (AUC), and results are summarized in Table \ref{table:experiment_metrics}. For each row, the mean area under the curve (AUC) score is reported based on 1000 bootstrap iterations. On average, 9,647 tiles were extracted at 20× magnification per slide (min: 46, median: 8,067, max: 45,172), depending on the size of the main tissue area.

\begin{table}[!ht]
    \centering
    \caption{Model performance across three dataset curations. AUC is one-vs-all, accuracy is balanced accuracy.}
    \scriptsize
    \begin{tabular}{cc|ccccc|cc}\toprule
    \multirow{2}{*}{Experiment} &\multirow{2}{*}{Encoder} &\multicolumn{5}{c}{AUCs by Racial Groups} &\multicolumn{2}{|c}{Overall Metrics} \\\cmidrule{3-9}
    & &White &Black &Hispanic &Asian &Other &AUC & Accuracy \\\midrule
    \multirow{2}{*}{Exp1} &SP22M &0.772 &0.785 &0.586 &0.805 &0.547 &0.699 &0.395 \\
    &UNI &0.797 &0.791 &0.607 &0.791 &0.603 &\textbf{0.718} &\textbf{0.400} \\
    Uncurated&GigaPath &0.801 &0.753 &0.598 &0.801 &0.522 &0.695 &0.388 \\
    &Virchow &0.784 &0.749 &0.591 &0.783 &0.579 &0.697 &0.392 \\
    &\underline{Average} &\textbf{0.789} &0.770 &0.596 &\textbf{\underline{0.795}} &0.563 &0.702 &0.394 \\ \cmidrule{1-9}
        \multirow{2}{*}{Exp2} &SP22M &0.744 &0.751 &0.569 &0.701 &0.577 &0.668 &0.368 \\
        &UNI &0.760 &0.773 &0.560 &0.715 &0.569 &\textbf{0.676} &\textbf{0.380} \\
        Balance&GigaPath &0.734 &0.739 &0.581 &0.753 &0.559 &0.673 &0.372 \\
        Disease&Virchow &0.728 &0.753 &0.529 &0.726 &0.590 &0.665 &0.334 \\
        &\underline{Average} &\textbf{0.742} &\textbf{0.754} &0.560 &\underline{0.724} &0.574 &0.671 &0.364 \\ \cmidrule{1-9}
        \multirow{2}{*}{Exp3} &SP22M &0.788 &0.773 &0.584 &0.481 &0.534 &0.632 &0.287 \\
        &UNI &0.819 &0.766 &0.654 &0.556 &0.594 &0.678 &0.296 \\
        Strict&GigaPath &0.791 &0.766 &0.664 &0.650 &0.431 &0.661 &\textbf{0.333} \\
        ICD code&Virchow &0.796 &0.742 &0.656 &0.592 &0.613 &\textbf{0.680} &0.293 \\
        &\underline{Average} &\textbf{0.799} &\textbf{0.762} &0.640 &\underline{0.570} &0.543 &0.663 &0.302 \\
    \bottomrule
    \end{tabular}
    \label{table:experiment_metrics}
\end{table}

\textbf{Exp1 (Uncurated)} included all available dermatopathology specimens and yielded the highest overall OvR AUC (0.702), with particularly strong performance in the Asian group (AUC = 0.795). This was attributed to a disproportionately high prevalence of hemorrhoid cases (61\%) among Asian patients due to site-specific sampling biases (160 out of 312 Asian patients treated at one site). \textbf{Exp2 (Balance Disease)} mitigated disease-related confounding by rebalancing hemorrhoid cases and removing gangrene and sun damage-related conditions disproportionately prevalent in Black and White patients but had low overall occurrence (e.g., melanoma, basal cell carcinoma, squamous cell carcinoma, actinic keratosis, and seborrheic keratosis), resulting in 2,032 patients (W 37.5\%, B 19.8\%, H/L 17.3\%, A 15.1\%, O 10.2\%). This adjustment led to a decline in overall OvR AUC (0.671), with the Asian group experiencing the largest drop (AUC: 0.795 → 0.724). In \textbf{Exp3 (Strict ICD Code)}, we further restricted the dataset to classical dermatopathology cases (ICD-10 code, L: inflammatory skin diseases, C: skin cancers, D: benign skin growths and disorders), fully removing hemorrhoids (ICD-10 K), and reducing dataset to 800 patients (W 46.9\%, B 19.9\%, H/L 19.6\%, A 7.2\%, O 6.5\%). This further reduced the overall OvR AUC to 0.663, with the Asian group showing the most pronounced decline (0.570), whereas the White group maintained consistently high performance (0.799).

\begin{figure*}[!ht]
    \centering
    \includegraphics[width=1.0\linewidth]{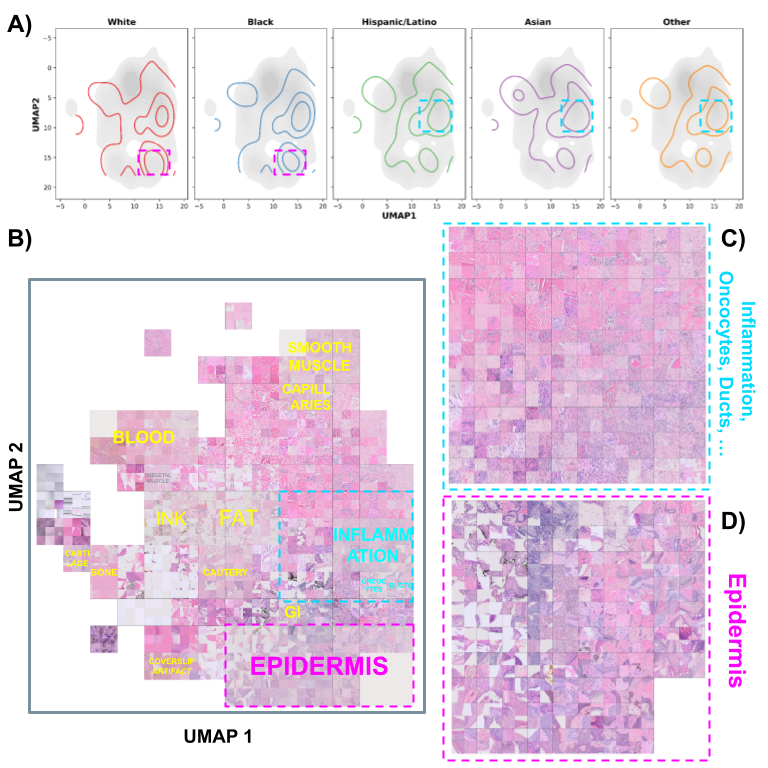}
    \caption{\textbf{UMAP visualization of attention scores.} (A) Density plot with a grayscale KDE background representing the overall distribution. Contour lines were generated for high-attention tiles within each racial group. (B) Grid plot visualizing representative samples from different UMAP regions. (C, D) Zoomed-in grid plots highlighting regions that received high model attention. GI: gastrointestinal tract.}
    \label{fig:umap_workflow}
\end{figure*}

\subsection{UMAP Visualization of Attention and Morphological Patterns}
Visual inspection of attention scores suggested a spatial association between high attention and tissue morphology across racial groups. To investigate this, we projected attention scores into a lower-dimensional UMAP space using SP22M encoder due to its lightweight architecture (22M parameters) and comparable performance. For UMAP generation, 20 slides per racial group were randomly sampled from \textbf{Exp3}, with 10,000 tiles per slide. Figure \ref{fig:umap_workflow}A presents the UMAP projection, where a grayscale kernel density estimation (KDE) background shows the overall data distribution, and colored contour lines highlight regions with the top 10\% of attention scores for each racial group. White and Black groups exhibit more concentrated attention clusters, whereas the Hispanic/Latino, Asian, and Other groups display more dispersed attention distributions, suggesting potential histomorphological differences. To further examine morphological differences in model attention, Figure \ref{fig:umap_workflow}B visualizes pathologist-annotated tissue types (epidermis, inflammation, blood, fat, background, etc.) from different UMAP regions, reinforcing that attention is influenced by distinct histological structures. Figure \ref{fig:umap_workflow}C and D zoom into two high-attention regions identified in the KDE plot. Figure \ref{fig:umap_workflow}C includes diverse histomorphological types (oncocytes, ducts, inflammation), but high attention in this region lacks a clear structural association. In contrast, Figure \ref{fig:umap_workflow}D corresponds specifically to epidermis, aligning with the strong epidermal attention observed in the White and Black groups. 

\begin{figure*}[!ht]
    \centering
    \includegraphics[width=1.0\linewidth]{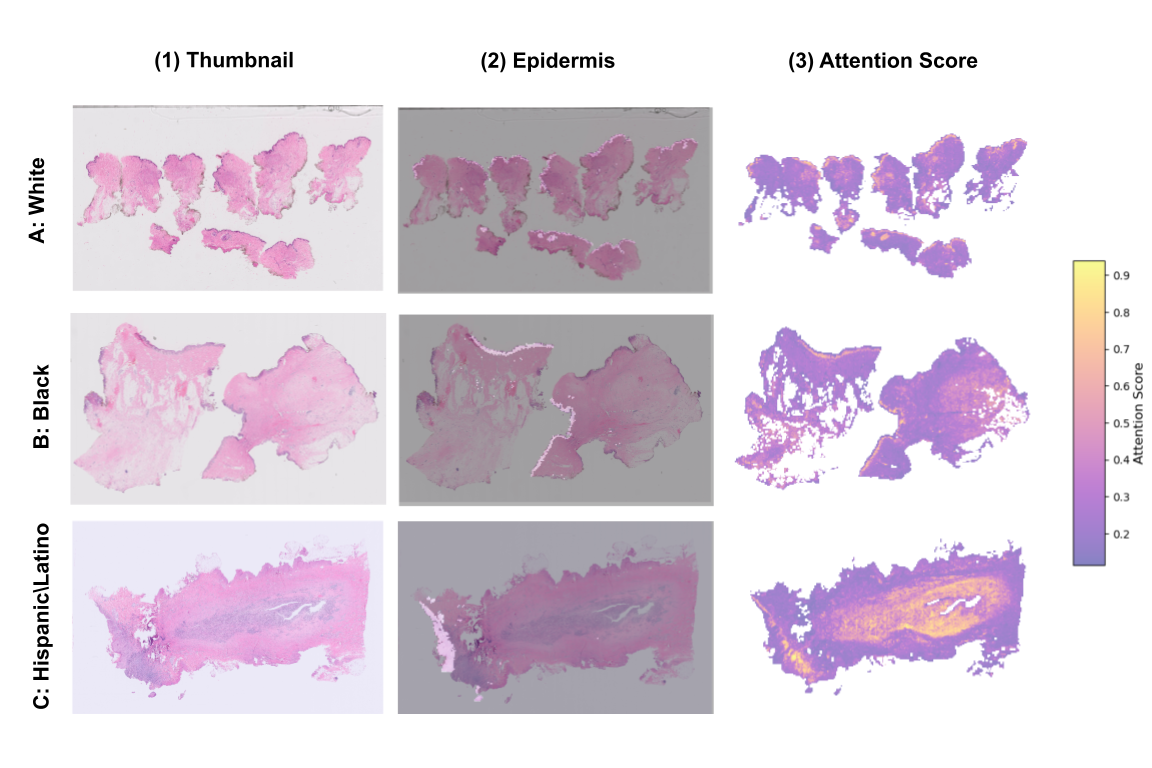}
    \caption{Whole-slide attention maps for selected examples from \textbf{Exp3}. Rows correspond to three selected slides, with columns showing: (1) Whole-slide thumbnail, (2) Binary mask of epidermis detection, and (3) Attention score from a specific racial group. (A) Attention to White, (B) Attention to Black, (C) Attention to Hispanic/Latino.}
    \label{fig:case_analysis}
\end{figure*}

\subsection{Attention Distribution Analysis}
\begin{figure*}[!ht]
    \centering
    \includegraphics[width=1\linewidth]{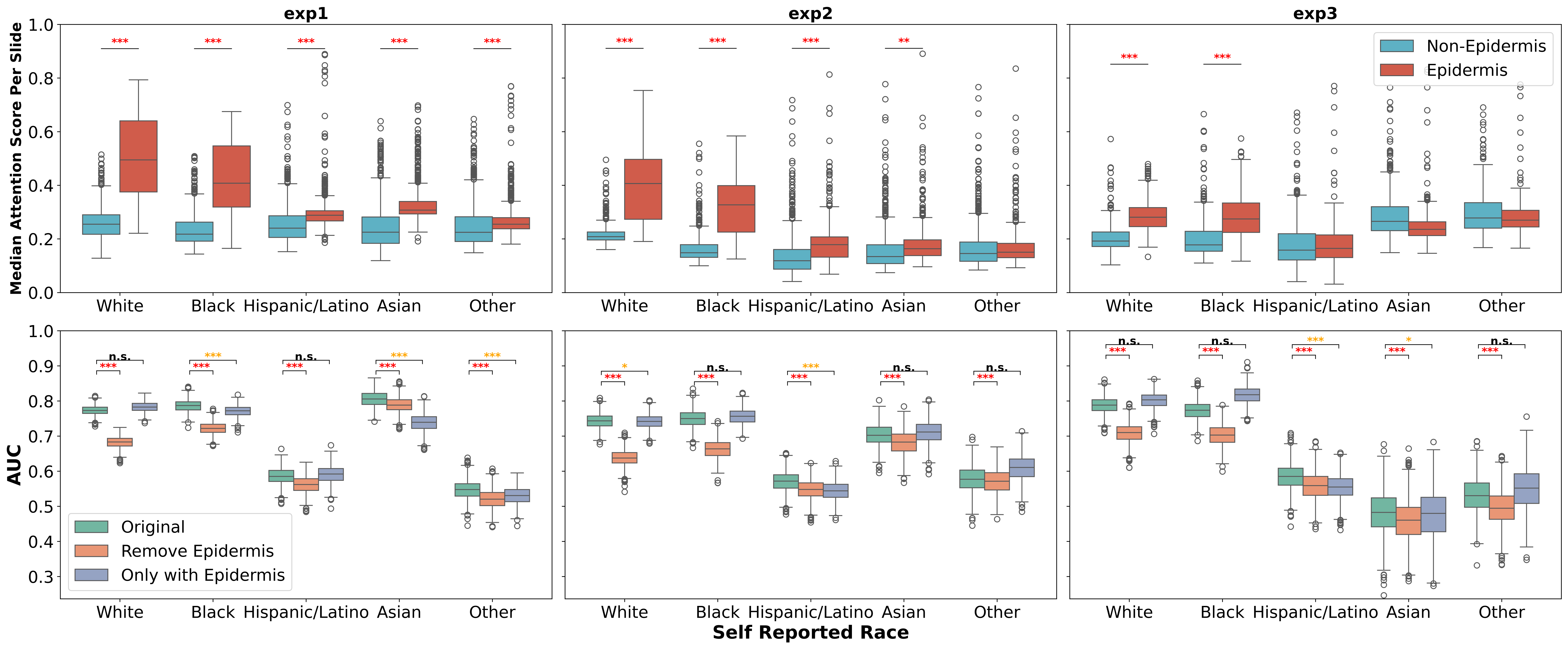}
    \caption{Attention and ablation analysis across racial groups and experiments. (A) Boxplots comparing the median attention score per slide between epidermis and non-epidermis regions (B) AUCs with epidermis tiles removed (orange), kept only (blue), and compared to original model (green). One-sided paired t-test significance: *: p<0.05, **: p<0.01, ***: p<0.0001, n.s./unlabeled if not significant.}
    \label{fig:box_plots}
\end{figure*}

Figure \ref{fig:case_analysis} presents whole-slide attention maps from examples in \textbf{Exp3}. In (A), attention to White maps strongly to the epidermis, whereas in (B), attention to Black highlights the epidermis but also extends to other regions. In (C), attention to Hispanic is predominantly observed in non-epidermis regions. To compare attention distribution, we performed a one-sided paired t-test to assess whether epidermal regions received higher attention. Figure \ref{fig:box_plots}A presents the median attention score per slide including only slides with more than 15 epidermis tiles were to reduce noise. Across all three experiments, epidermal regions consistently received higher attention than non-epidermis regions, but the magnitude of this difference decreased from \textbf{Exp1} to \textbf{Exp3}. In \textbf{Exp3}, the effect was significant only for the White and Black groups, while the Asian group exhibited lower attention in epidermis than non-epidermis regions. Figure \ref{fig:box_plots}B further highlights the importance of epidermis in self-reported race prediction. When epidermis tiles were completely removed in validation data, model performance dropped by approximately 0.05 across all racial groups and experiments. When only epidermis tiles were retained—with 85\% of slides in validation set containing less than 20\% epidermis tiles—the model maintained comparable or even improved performance in some racial groups.

\section{Discussion}
In this study, we examined whether DL models can predict self-reported race from digitized dermatopathology slides, independent of pathology task. Unlike previous studies on reporting the differential performance of task-specific models, we explored whether biological correlates of race could be identified in histology images. This is important because task-specific models could exploit these features as shortcuts, leading to unintended biases and disparities in clinical predictions. Our results (\textbf{Exp3}) show that self-reported race can be predicted with moderate accuracy (AUC = 0.7), particularly for White (0.80) and Black (0.76) patients. Across four encoders used, UNI consistently captured race-related information, despite Virchow being pretrained on the largest number of skin slides (18\%, 273,893 slides). 

We identified epidermis, which typically constitute 10\%–20\% of tissue in skin histology slides, as the strongest predictive histological component, consistent with the role of melanocytes in skin tone \cite{williams2023skin}. Across all experiments, White and Black groups consistently had higher prediction performance than Hispanic/Latino and Asian groups, possibly due to more distinct epidermal features in White and Black patients, whereas Hispanic and Asian groups exhibit greater morphological variation, making classification more challenging. Confounding variables in patient sampling and disease presentation also inflated prediction performance. In \textbf{Exp1}, the overrepresentation of hemorrhoid cases in Asian patients (61\%) likely caused the model to associate race with disease prevalence rather than intrinsic histological differences. \textbf{Exp2} rebalanced disease distribution, resulting in a performance drop, suggesting that race labels acted as unintended shortcuts for disease classification. \textbf{Exp3}, which applied ICD-10 coding to focus on skin disease cases, was chosen for further investigation.

% We selected Exp3 (Strict ICD Code) for further investigation because it isolates skin disease cases based on ICD-10 codes, reducing bias from non-skin conditions. Although some gastrointestinal (GI) tiles remained in Exp3, their impact was minimal. 

Our study raises key concerns about demographic shortcuts in CPath but also has several limitations. Self-reported race is a socioeconomic determinant that while identifying potential confounders, also introduces noise. The heterogeneity of the Hispanic/Latino group further complicates isolating specific biological or morphological patterns. Integrating genetic ancestry data alongside self-reported race may provide a more comprehensive understanding of demographic influences in histological analysis. This study focused on skin histology for better control over confounders, but future work should extend to other specimen/organs to determine whether race-associated patterns emerge in other tissues or if skin remains unique due to its link to pigmentation. Additionally, ICD-10 coding has inherent limitations, as it reflects clinical suspicion rather than definitive histological diagnosis. In addition, while we focused on removing high-attention epidermal tiles during validation, further investigation into secondary high-attention regions with more heterogenous morphologies would be valuable. Furthermore, our study used AB-MIL, a spatially-unaware aggregation model, meaning it analyzes each tile independently without considering spatial interactions across the slide. Evaluating on a more sophisticated slide-level aggregator that accounts for tile-to-tile spatial relationships could offer deeper insights into how histological structures collectively contribute to racial classification.

\section{conclusion}
While histological images may not encode demographic signals as strongly as radiological images \cite{gichoya2022ai,adleberg2022predicting}, DL models can still predict self-reported race, likely by leveraging morphological shortcuts such as epidermal structures in skin slides. These findings highlight the need to consider demographic biases in CPath models and the impact of dataset curation on model fairness. Developing bias mitigation methods to address model reliance on demographic shortcuts is crucial for ensuring fairness in CPath applications, and we encourage researchers to carefully account for disease distribution and remain mindful of how AI models may inadvertently learn and exploit sensitive demographic information rather than focusing on disease-related histological features.

\section*{Acknowledgments}
This work is supported in part through the use of research platform AI-Ready Mount Sinai (AIR.MS) and the expertise provided by the team at the Hasso Plattner Institute for Digital Health at Mount Sinai (HPI.MS). This work was supported in part through the computational and data resources and staff expertise provided by Scientific Computing and Data at the Icahn School of Medicine at Mount Sinai and supported by the Clinical and Translational Science Awards (CTSA) grant UL1TR004419 from the National Center for Advancing Translational Sciences. 

%Bibliography
\bibliographystyle{unsrt}  
\bibliography{references}

\end{document}